\pdfoutput=1

\documentclass[11pt]{article}

\usepackage[]{ACL2023}

\usepackage{times}
\usepackage{latexsym}

\usepackage[T1]{fontenc}
\usepackage[T5]{fontenc}

\usepackage[utf8]{inputenc}

\usepackage{microtype}

\usepackage{inconsolata}

\usepackage{amsthm}
\usepackage{amssymb}
\usepackage{tipa}
\usepackage{ragged2e}

\usepackage{hyperref}       
\usepackage{url}            
\usepackage{booktabs}       
\usepackage{amsfonts}       
\usepackage{amsmath}
\usepackage{nicefrac}       
\usepackage{microtype}      
\usepackage{lipsum}
\usepackage{graphicx}
\graphicspath{{figure/}}     
\usepackage{caption}
\usepackage{subcaption}
\usepackage{multirow}
\usepackage{booktabs}

\usepackage[final]{pdfpages}
\usepackage{tabularx}
\usepackage{tkz-tab}
\usepackage{framed,multicol}
\usepackage[framemethod=tikz]{mdframed}
\usepackage{xcolor}

\usepackage{wasysym}

\usepackage{pgfplots}
\usepackage{tikz,pgfplots,pgfplotstable}

\usepackage{framed,multicol}
\usepackage[framemethod=tikz]{mdframed}

\usepgfplotslibrary{colorbrewer}
\pgfplotsset{compat = 1.15, cycle list/Set1-8} 
\usetikzlibrary{pgfplots.statistics, pgfplots.colorbrewer} 
\usepackage{pgfplotstable}
\usepackage{filecontents}

\definecolor{markercolor}{RGB}{124.9, 255, 160.65}
\pgfplotsset{compat=1.3}
\usetikzlibrary{positioning, fit, calc}  
\tikzset{block/.style={draw, thick, text width=2cm ,minimum height=1.3cm, align=center},   
	line/.style={-latex}     
} 

\tikzset{blocktext/.style={draw, thick, text width=5.2cm ,minimum height=1.3cm, align=center},   
	line/.style={-latex}     
} 

\title{ChatGPT is Good but Bing Chat is Better for Vietnamese Students}

\author{Xuan-Quy Dao \\
	Eastern International University \\
	\texttt{quy.dao@eiu.edu.vn} 
	\And
	Ngoc-Bich Le \\
	International University Vietnam \\
	\texttt{lnbich@hcmiu.edu.vn} 
}

\begin{document}
	
	\maketitle
	
	\begin{abstract}
		
		This study examines the efficacy of two SOTA large language models (LLMs), namely ChatGPT and Microsoft Bing Chat (BingChat), in catering to the needs of Vietnamese students. Although ChatGPT exhibits proficiency in multiple disciplines, Bing Chat emerges as the more advantageous option. We conduct a comparative analysis of their academic achievements in various disciplines, encompassing mathematics, literature, English language, physics, chemistry, biology, history, geography, and civic education. The results of our study suggest that BingChat demonstrates superior performance compared to ChatGPT across a wide range of subjects, with the exception of literature, where ChatGPT exhibits better performance. Additionally, BingChat utilizes the more advanced GPT-4 technology in contrast to ChatGPT, which is built upon GPT-3.5. This allows BingChat to improve to comprehension, reasoning and generation of creative and informative text. Moreover, the fact that BingChat is accessible in Vietnam and its integration of hyperlinks and citations within responses serve to reinforce its superiority. In our analysis, it is evident that while ChatGPT exhibits praiseworthy qualities, BingChat presents a more adapted solutions for Vietnamese students.
		
	\end{abstract}
	
	\section{Introduction}
	
	Recent advancements in LLMs have enabled AI to understand and communicate with humans, creating opportunities for its use in education. LLMs have shown great potential in education, content development, and language translation. The two primary architectures of LLMs are BERT (Bidirectional Encoder Representations from Transformers) and GPT (Generative Pre-trained Transformer). In 2018, Google introduced BERT~\cite{devlin2018bert}, which excelled in various natural language processing (NLP) tasks. OpenAI developed the GPT algorithm~\cite{alec2018improving}, which was trained on extensive unlabeled text datasets. Facebook’s RoBERTa~\cite{liu2019roberta} built on Google’s research, and Google released T5~\cite{raffel2020exploring} in 2019. In 2020, OpenAI created GPT-3~\cite{brown2020language}, which demonstrated exceptional performance in various NLP tasks. Recently, OpenAI developed GPT-4~\cite{OpenAI_gpt_4_report}, a text-to-text machine learning system capable of processing both text and image inputs. GPT-4 has shown human-level performance in many professional and academic criteria, although it may not perform as well as humans in other contexts.
	
	LLMs need to provide citations and hyperlinks in generated answers in order to become useful tool in education. Liu et al.~\cite{liu2023evaluating} proposed ALCE, a benchmark for evaluating the citation quality of text generated by LLMs. Their experiments reveal room for improvement, as even the best LLM model lacks complete citation support in 49\% of its generations on the ELI5 dataset. Gao et al.~\cite{gao2023enabling} discussed the importance of verifiability in generative search engines and find that existing engines often provide fluent but unsupported and inaccurately cited responses. They highlight the low rates of fully supported sentences (51.5\%) and supporting citations (74.5\%), emphasizing the need for trustworthy information sources.
	
	In~\cite{dao2023exa}, ChatGPT can pass the Vietnamese National High School Graduation Examination (VNHSGE).  Nevertheless, ChatGPT is currently unavailable in Vietnam, whereas BingChat is accessible. The objective of this study is to conduct a comparative analysis of the efficacy of ChatGPT and BingChat in providing academic support to Vietnamese students across multiple academic fields. To comprehensively evaluate the capabilities of both models, it is important to acknowledge that ChatGPT utilizes GPT-3.5 technology, while BingChat utilizes the more advanced GPT-4 technology. This distinction presents potential benefits in terms of LLMs' ability to understand intricate queries and generate text that is both creative and informative.  Consequently, we adhere to the formulation of research questions::
	\begin{itemize}
		\item RS1: What is BingChat's performance on the VNHSGE examination?
		\item RS2: How does BingChat's ability compare to ChatGPT's ability on the VNHSGE examination?
		\item RS3: How does BingChat's ability compare to Vietnamese students' ability on the VNHSGE examination?
		\item RS4: Can BingChat replace ChatGPT in education in Vietnam?
	\end{itemize}
	
	This research constitutes a contribution by demonstrating BingChat's effective performance in the VNHSGE examination. It emphasizes the superiority of BingChat over ChatGPT in the context of Vietnamese education. Additionally, the paper compares BingChat's performance to that of Vietnamese students across various subjects. Furthermore, the availability of BingChat as a feasible educational tool in Vietnam is discussed. This paper presents the following contributions: 
	
	\begin{itemize}
		
	\item The study presents evidence of BingChat's effective performance in the context of the VNHSGE examination. The discovery holds considerable importance, as it demonstrates the capability of BingChat to successfully navigate a rigorous examination that aims to evaluate students' comprehension across diverse disciplines. This implies that BingChat may hold significant value as a tool for students who are preparing for the VNHSGE examination, or for students seeking LLMs-based assistant tools in the field of education.
	
	\item The study emphasizes that BingChat demonstrates a higher level of performance in comparison to ChatGPT within the specific context of Vietnamese education. This finding holds significance as it implies that BingChat possesses distinct advantages over ChatGPT that are particularly pertinent to the Vietnamese education system. BingChat demonstrates superior capabilities in accessing and processing information derived from Vietnamese sources, while also offering the added functionality of providing hyperlinks within its responses.
	
	\item The study demonstrates that BingChat exhibits superior academic performance compared to Vietnamese students in various subjects, including English, literature, biology, history, and geography. However, BingChat's performance in civic education is found to be comparable to that of Vietnamese students. Nevertheless, the data indicates that Vietnamese students exhibit superior performance compared to BingChat in the fields of mathematics, physics, and chemistry. The aforementioned discovery is of particular interest, as it implies that BingChat exhibits varying levels of proficiency across different subject matters. There are several potential factors that may contribute to this phenomenon, including the training methodology employed for BingChat and the inherent characteristics of the VNHSGE examination. BingChat currently lacks the necessary capabilities to effectively serve as an educational tool in subjects that demand sophisticated cognitive processes, such as mathematics, physics, and chemistry.
	
	\item This paper examines the viability of BingChat as a substitute for ChatGPT within the Vietnamese education system, given BingChat's superior capabilities and the absence of official availability of ChatGPT in Vietnam. This final point holds significance as it implies that the integration of BingChat into the Vietnamese education system could yield substantial benefits, offering students a valuable resource for educational purposes and academic endeavors.

	\end{itemize}
	
	\section{Related works}
	
	\subsection{Evaluation of LLMs on education}
	
	Previous research has investigated multiple facets pertaining to the utilization of ChatGPT, a prominent language model, within the realm of education. GPT-4~\cite{OpenAI_gpt_4_report} demonstrates a level of performance comparable to that of humans across a range of professional and academic assessments, such as Uniform Bar Exam, AP Chemistry, GRE Quantitative section, and AP Physics 2. In contrast, the performance of GPT-3.5 on these benchmarks is notably inferior compared to that of GPT-4

	In previous studies explored various aspects of using ChatGPT, a large language model, in the field of education. GPT-4~\cite{OpenAI_gpt_4_report} exhibits human-level performance on a variety of professional and academic benchmarks, including the Uniform Bar Exam, AP Chemistry, the GRE Quantitative section, and AP Physics 2. In contrast, GPT-3.5 does not perform as well on these benchmarks, scoring significantly lower than GPT-4.  In higher education~\cite{rudolph2023war}, GPT-4 demonstrates superior performance, whereas BingChat and Bard exhibit comparatively lower performance, resembling that of academically challenged students with failing grades.

	LLMs have the potential in supporting medical education and clinical decision-making~\cite{kung2023performance}, answering medical questions~\cite{gilson2023does}, supporting learning and assessment in law exams~\cite{Choi2022}, assisting academic paper writing and suggests a shift in learning goals~\cite{Zhai2022}. 
	
	Prior research has also investigated the benefits, limitations, and challenges related to the adoption of LLMs in the realm of education. It underscores the necessity of well-defined strategies and the cultivation of critical thinking skills~\cite{kasneci2023chatgpt}. Additionally, it highlights the significance of ethical considerations in the context of teaching and learning~\cite{baidoo2023education}, particularly in relation to personalized learning and virtual simulations~\cite{qadir2023engineering}. The study also addresses concerns regarding academic dishonesty and breaches of privacy~\cite{tlili2023if}, emphasizing the importance of safeguarding privacy rights, promoting fairness, and ensuring transparency~\cite{mhlanga2023open}.
	
	
	\subsection{Evaluation of LLMs on Vietnamese education}

	In their study, Quy et al. conducted an evaluation of the performance of LLMs on the VNHSGE examination~\cite{dao2023vnhsge, dao2023math, dao2023che, dao2023phy, dao2023eng, dao2023exa}. In order to assess the efficacy of LLMs in the context of the VNHSGE examination, Quy et al. constructed a dataset known as VNHSGE dataset~\cite{dao2023vnhsge}. This dataset encompasses nine distinct subjects and was derived from the VNHSGE examination as well as other comparable examinations. The dataset comprises a total of 300 literary essays and 19,000 multiple-choice questions pertaining to various subjects. The dataset possesses the potential to assess the performance of LLMs across various multitask scenarios, including but not limited to question answering, text generation, reading comprehension, visual question answering, and other related tasks.

	The recent research conducted a thorough evaluation of ChatGPT's mathematical proficiency by assessing its performance in answering multiple-choice questions from the VNHSGE examination, covering various subjects and levels of difficulty~\cite{dao2023math}. The findings indicated that ChatGPT exhibited varying levels of proficiency in mathematics, which were contingent upon the complexity of the tasks and the specific subject matter. The system demonstrated superior performance on knowledge-level questions, achieving an accuracy rate of 83\%. Nevertheless, as the complexity of the task increased, there was a corresponding decline in the accuracy rate. Specifically, the scores for comprehension level, application level, and high application level were recorded at 62\%, 27\%, and 10\% respectively. In a related study, the authors explored the performance of ChatGPT and BingChat in various domains, as documented in references~\cite{dao2023che, dao2023phy, dao2023eng, dao2023vnhsge}.

	According to Dao et al.~\cite{dao2023exa}, ChatGPT achieved a mean score ranging from 6 to 7 on the VNHSGE, thereby showcasing the transformative capabilities of this technology in reshaping the educational domain. The evaluation of ChatGPT's performance unveiled its competence across various disciplines such as mathematics, English, physics, chemistry, biology, history, geography, civic education, and literature. This implies that it has the capability to offer efficient assistance to learners.

	\subsection{Hyperlink Feature in Answers}
	
	\subsubsection{Importance of Citations and Hyperlinks in LLMs-Generated Answers}
	
	The utilization of LLMs is experiencing a growing trend in various domains, encompassing education, research, and entertainment. Nevertheless, it is crucial to ascertain the accuracy and reliability of the information presented by LLMs. Specifically, citations and hyperlinks play a crucial role in academic discourse.
	
	\textbf{Citations}: Citations play a crucial role in guaranteeing the veracity of the responses provided in LLMs. By citing authoritative references, LLMs can substantiate their assertions with empirical data, thereby guaranteeing the credibility and dependability of the information presented. Citations additionally foster transparency and accountability by affording users the opportunity to scrutinize the sources of information utilized by LLMs. This feature facilitates users in making informed assessments regarding the veracity and dependability of the content that is generated.
	
	\textbf{Hyperlinks}: Hyperlinks are an additional essential element for guaranteeing the quality of LLMs’ responses. By supplying hyperlinks to the original sources, LLMs allow users to directly access the information and validate the model's claims. This encourages critical thinking and a culture of fact-checking. In addition, hyperlinks enable users to investigate related information beyond the generated response, which can be beneficial for further study and learning.
	
	\subsubsection{Benefits of Citations and Hyperlinks}
	
	The incorporation of citations and hyperlinks within LLMs responses can yield several advantages, such as:
	\begin{itemize}
		\item \textbf{Factual accuracy}: Citations play a crucial role in guaranteeing the accuracy and reliability of information provided by LLMs. 
		\item \textbf{Transparency}: Citations play a crucial role in fostering transparency by disclosing the origins of information utilized by LLMs.
		\item \textbf{Verifiability}: Hyperlinks can serve as a means to enable autonomous verification of the information provided by LLMs.
		\item \textbf{Further exploration}: Hyperlinks can facilitate users in accessing supplementary information that is pertinent to the generated response.
		\item \textbf{Learning and research}: Citations and hyperlinks are of significant value to students, researchers, and individuals who aim to enhance their comprehension of a particular subject.
	\end{itemize}
	
	Overall, citations and hyperlinks in LLMs answers contribute to factual accuracy, transparency, verifiability, and facilitate further learning and research. By incorporating these features, LLMs can enhance the quality and reliability of the information they provide.
	
		\begingroup
	\renewcommand{\arraystretch}{1.5} 
	\begin{table*}[ht!]
		\caption{ChatGPT and BingChat performances on VNHSGE dataset~\cite{dao2023vnhsge}}
		\label{tabl:llm_result}
		\resizebox{\textwidth}{!}{%
			\begin{tabular}{c|cc|cc|cc|cc|cc|cc|cc|cc|cc|}
				\cline{2-19}
				& \multicolumn{2}{c|}{\textbf{Math}}               & \multicolumn{2}{c|}{\textbf{Lit}}              & \multicolumn{2}{c|}{\textbf{Eng}}                  & \multicolumn{2}{c|}{\textbf{Phy}}              & \multicolumn{2}{c|}{\textbf{Che}}                & \multicolumn{2}{c|}{\textbf{Bio}}              & \multicolumn{2}{c|}{\textbf{His}}                  & \multicolumn{2}{c|}{\textbf{Geo}}                  & \multicolumn{2}{c|}{\textbf{Civ}}                  \\ \hline
				\multicolumn{1}{|c|}{\textbf{2019}} & \multicolumn{1}{c|}{52}            & 56          & \multicolumn{1}{c|}{75}          & 52.75          & \multicolumn{1}{c|}{76}            & 92            & \multicolumn{1}{c|}{60}          & 55          & \multicolumn{1}{c|}{40}          & 55            & \multicolumn{1}{c|}{60}          & 67.5        & \multicolumn{1}{c|}{42.5}          & 82.5          & \multicolumn{1}{c|}{50}            & 75            & \multicolumn{1}{c|}{60}            & 75            \\ \hline
				\multicolumn{1}{|c|}{\textbf{2020}} & \multicolumn{1}{c|}{66}            & 56          & \multicolumn{1}{c|}{68.9}        & 51.25        & \multicolumn{1}{c|}{86}            & 96            & \multicolumn{1}{c|}{62.5}        & 67.5        & \multicolumn{1}{c|}{42.5}        & 57.5          & \multicolumn{1}{c|}{60}          & 72.5        & \multicolumn{1}{c|}{47.5}          & 85            & \multicolumn{1}{c|}{52.5}          & 70            & \multicolumn{1}{c|}{70}            & 87.5          \\ \hline
				\multicolumn{1}{|c|}{\textbf{2021}} & \multicolumn{1}{c|}{60}            & 66          & \multicolumn{1}{c|}{75}          & 60.25          & \multicolumn{1}{c|}{76}            & 86            & \multicolumn{1}{c|}{60}          & 67.5        & \multicolumn{1}{c|}{62.5}        & 50            & \multicolumn{1}{c|}{52.5}        & 67.5        & \multicolumn{1}{c|}{55}            & 90            & \multicolumn{1}{c|}{75}            & 82.5          & \multicolumn{1}{c|}{62.5}          & 92.5          \\ \hline
				\multicolumn{1}{|c|}{\textbf{2022}} & \multicolumn{1}{c|}{62}            & 60          & \multicolumn{1}{c|}{56.3}        & 70        & \multicolumn{1}{c|}{80}            & 94            & \multicolumn{1}{c|}{65}          & 67.5        & \multicolumn{1}{c|}{47.5}        & 47.5          & \multicolumn{1}{c|}{57.5}        & 72.5        & \multicolumn{1}{c|}{60}            & 92.5          & \multicolumn{1}{c|}{62.5}          & 85            & \multicolumn{1}{c|}{82.5}          & 90            \\ \hline
				\multicolumn{1}{|c|}{\textbf{2023}} & \multicolumn{1}{c|}{54}            & 62          & \multicolumn{1}{c|}{64.8}        & 49.75        & \multicolumn{1}{c|}{78}            & 94            & \multicolumn{1}{c|}{57.5}        & 72.5        & \multicolumn{1}{c|}{47.5}        & 52.5          & \multicolumn{1}{c|}{60}          & 65          & \multicolumn{1}{c|}{77.5}          & 92.5          & \multicolumn{1}{c|}{67.5}          & 85            & \multicolumn{1}{c|}{77.5}          & 82.5          \\ \hline
				\multicolumn{1}{|c|}{\textbf{AVG}}  & \multicolumn{1}{c|}{\textbf{58.8}} & \textbf{60} & \multicolumn{1}{c|}{\textbf{68}} & \textbf{56.8} & \multicolumn{1}{c|}{\textbf{79.2}} & \textbf{92.4} & \multicolumn{1}{c|}{\textbf{61}} & \textbf{66} & \multicolumn{1}{c|}{\textbf{48}} & \textbf{52.8} & \multicolumn{1}{c|}{\textbf{58}} & \textbf{69} & \multicolumn{1}{c|}{\textbf{56.5}} & \textbf{88.5} & \multicolumn{1}{c|}{\textbf{61.5}} & \textbf{79.5} & \multicolumn{1}{c|}{\textbf{70.5}} & \textbf{85.5} \\ \hline
			\end{tabular}
		}
	\end{table*}
	\endgroup
	
		\begin{figure*}[h!]
		\begin{center}
			\begin{tikzpicture}  
				
				\begin{axis}  
					[  
					ybar, 
					bar width=15pt, 
					legend style={at={(0.5,-0.35)}, 	
						anchor=north,legend columns=-1},    
					symbolic x coords={Mathematics, Literature, English, Physics, Chemistry, Biology, History, Geography, Civic Education}, 
					xtick=data,
					x tick label style={rotate=45,anchor=east},
					yticklabel={\pgfmathprintnumber{\tick}\%},  
					nodes near coords,   
					every node near coord/.append style={rotate=90, anchor=west},
					enlarge x limits,
					height=6cm, width=16cm,
					axis x line*=bottom, axis y line*=left 
					]  
					\addplot [fill=yellow] coordinates {
						(Mathematics, 58.8)
						(Literature, 68)
						(English, 79.2)
						(Physics, 61)
						(Chemistry, 48)
						(Biology, 58)
						(History,56.5)
						(Geography, 61.5)
						(Civic Education, 70.5)
						
					}; 
					\addplot [fill=red] coordinates {
						(Mathematics, 60)
						(Literature, 56.8)
						(English, 92.4)
						(Physics, 66)
						(Chemistry, 52.5)
						(Biology, 69)
						(History,88.5)
						(Geography, 79.5)
						(Civic Education, 85.5)
						
					}; 
					\legend{
						ChatGPT,
						BingChat,}
					
				\end{axis}  
			\end{tikzpicture}
		\end{center}
		\caption{Comparison of ChatGPT and BingChat performances on VNHSGE dataset (Figure from~\cite{dao2023vnhsge}).}
		\label{fig:ChatGPT_BingChat_exam}
	\end{figure*}
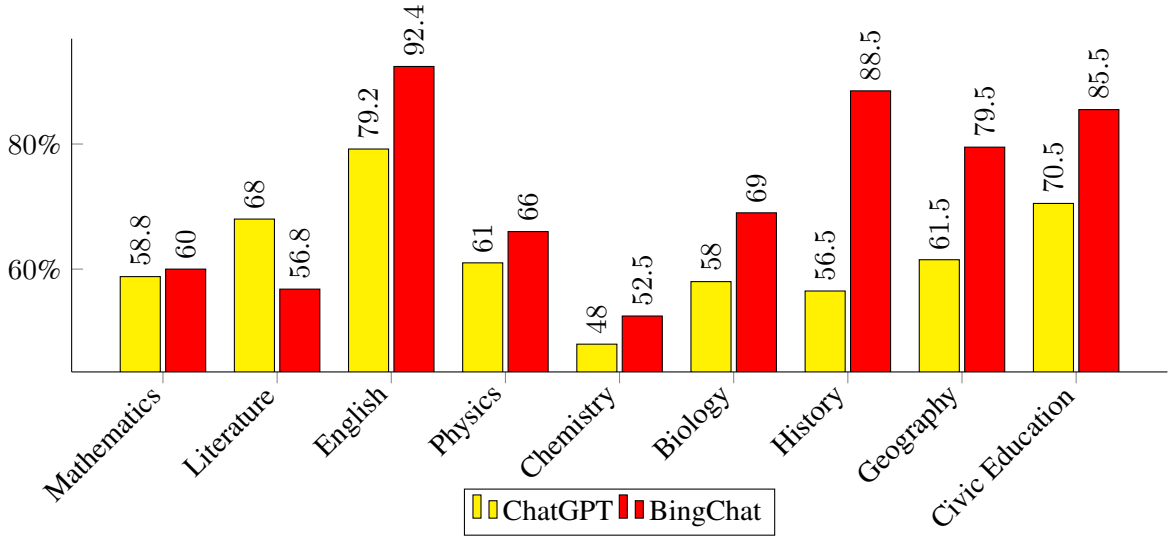
	
	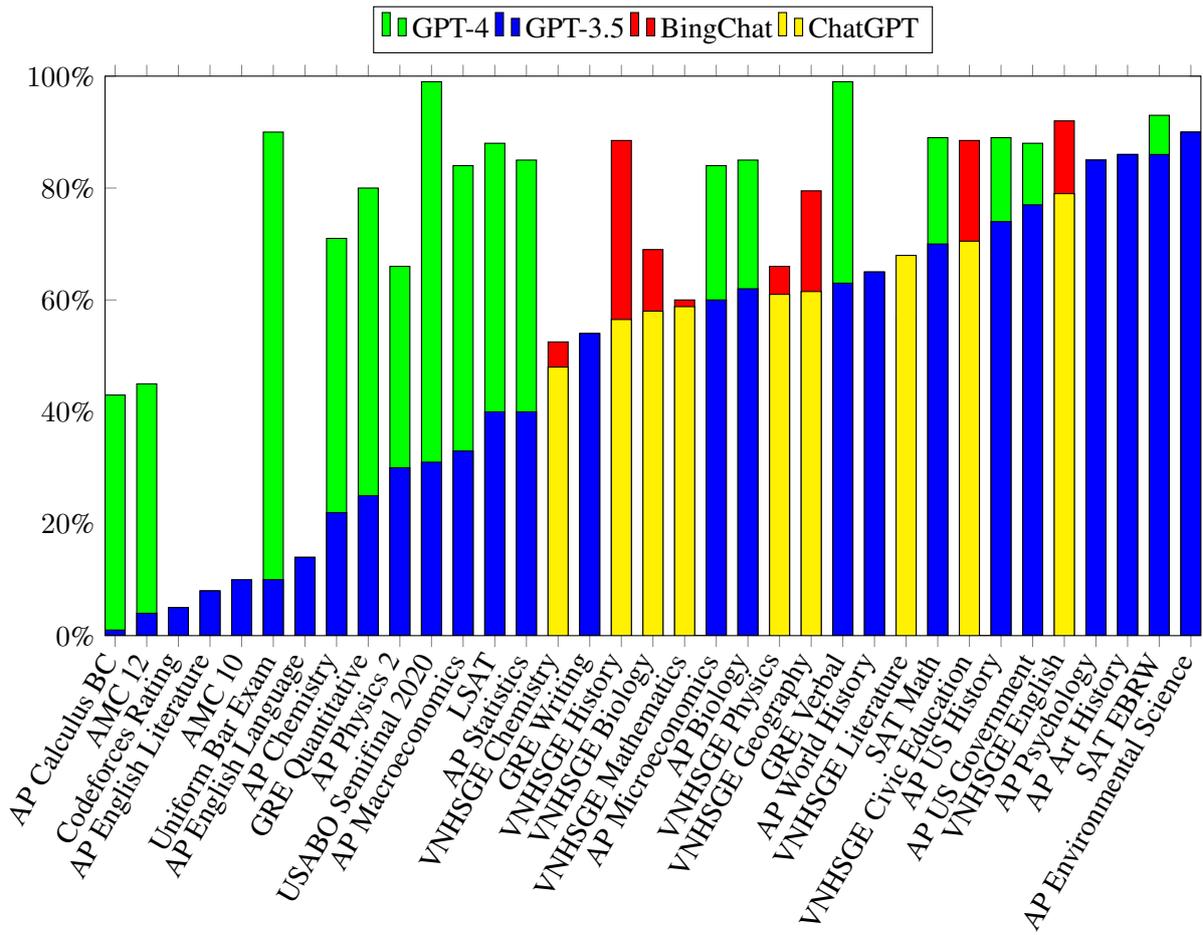
\begin{figure*}[h!]
		\begin{center}
			\begin{tikzpicture}
				\begin{axis}[
					ylabel={\pgfmathprintnumber{\tick}\%},
					legend style={at={(0.5,1.125)}, 	
						anchor=north,legend columns=-1},
					symbolic x coords={
						AP Calculus BC,
						AMC 12,
						Codeforces Rating,
						AP English Literature,
						AMC 10,
						Uniform Bar Exam,
						AP English Language,
						AP Chemistry,
						GRE Quantitative,
						AP Physics 2,
						USABO Semifinal 2020,
						AP Macroeconomics,
						LSAT,
						AP Statistics,
						VNHSGE Chemistry,
						GRE Writing,
						VNHSGE History,
						VNHSGE Biology,
						VNHSGE Mathematics,
						AP Microeconomics,
						AP Biology,
						VNHSGE Physics,
						VNHSGE Geography,
						GRE Verbal,
						AP World History,
						VNHSGE Literature,
						SAT Math,
						VNHSGE Civic Education,
						AP US History,
						AP US Government,
						VNHSGE English,
						AP Psychology,
						AP Art History,
						SAT EBRW,
						AP Environmental Science,
					},
					xtick=data,
					hide axis,
					x tick label style={rotate=60,anchor=east},
					ybar,
					bar width=7.5pt,
					ymin=0,
					ymax=100,
					width=\textwidth, 
					enlarge x limits={abs=0.5*\pgfplotbarwidth},					
					height=9cm, width=16cm,
					]
					\addplot [fill=green] coordinates {
						(AP Calculus BC,0)	
					};
					\addplot [fill=blue] coordinates {
						(AP Microeconomics,0)
					};	
					\addplot [fill=red] coordinates {
						(VNHSGE Civic Education,0)	
					};	
					\addplot [fill=yellow] coordinates {
						(AP Environmental Science,0)	
					};		
					\legend{GPT-4, GPT-3.5, BingChat, ChatGPT}				
				\end{axis}
				\begin{axis}[
					ylabel={},
					symbolic x coords={
						AP Calculus BC,
						AMC 12,
						Codeforces Rating,
						AP English Literature,
						AMC 10,
						Uniform Bar Exam,
						AP English Language,
						AP Chemistry,
						GRE Quantitative,
						AP Physics 2,
						USABO Semifinal 2020,
						AP Macroeconomics,
						LSAT,
						AP Statistics,
						VNHSGE Chemistry,
						GRE Writing,
						VNHSGE History,
						VNHSGE Biology,
						VNHSGE Mathematics,
						AP Microeconomics,
						AP Biology,
						VNHSGE Physics,
						VNHSGE Geography,
						GRE Verbal,
						AP World History,
						VNHSGE Literature,
						SAT Math,
						VNHSGE Civic Education,
						AP US History,
						AP US Government,
						VNHSGE English,
						AP Psychology,
						AP Art History,
						SAT EBRW,
						AP Environmental Science,
					},
					xtick=data,
					x tick label style={rotate=60,anchor=east},
					yticklabel={\pgfmathprintnumber{\tick}\%},
					ybar,
					bar width=7.5pt,
					ymin=0,
					ymax=100,
					width=\textwidth, 
					enlarge x limits={abs=0.5*\pgfplotbarwidth},					
					height=9cm, width=16cm,
					]
					\addplot [fill=green] coordinates {
						(AP Calculus BC,43)
						(AMC 12,45)
						(Codeforces Rating,5)
						(AP English Literature,8)
						(AMC 10,6)
						(Uniform Bar Exam,90)
						(AP English Language,14)
						(AP Chemistry,71)
						(GRE Quantitative,80)
						(AP Physics 2,66)
						(USABO Semifinal 2020,99)
						(AP Macroeconomics,84)
						(LSAT,88)
						(AP Statistics,85)
						(VNHSGE Chemistry,0)
						(GRE Writing,54)
						(VNHSGE History,0)
						(VNHSGE Biology,0)
						(VNHSGE Mathematics,0)
						(AP Microeconomics,84)
						(AP Biology,85)
						(VNHSGE Physics,0)
						(VNHSGE Geography,0)
						(GRE Verbal,99)
						(AP World History,65)
						(VNHSGE Literature,0)
						(SAT Math,89)
						(VNHSGE Civic Education,0)
						(AP US History,89)
						(AP US Government,88)
						(VNHSGE English,0)
						(AP Psychology,85)
						(AP Art History,86)
						(SAT EBRW,93)
						(AP Environmental Science,90)	
					};					
				\end{axis}
				\begin{axis}[
					ylabel={Percentage},
					symbolic x coords={
						AP Calculus BC,
						AMC 12,
						Codeforces Rating,
						AP English Literature,
						AMC 10,
						Uniform Bar Exam,
						AP English Language,
						AP Chemistry,
						GRE Quantitative,
						AP Physics 2,
						USABO Semifinal 2020,
						AP Macroeconomics,
						LSAT,
						AP Statistics,
						VNHSGE Chemistry,
						GRE Writing,
						VNHSGE History,
						VNHSGE Biology,
						VNHSGE Mathematics,
						AP Microeconomics,
						AP Biology,
						VNHSGE Physics,
						VNHSGE Geography,
						GRE Verbal,
						AP World History,
						VNHSGE Literature,
						SAT Math,
						VNHSGE Civic Education,
						AP US History,
						AP US Government,
						VNHSGE English,
						AP Psychology,
						AP Art History,
						SAT EBRW,
						AP Environmental Science,
					},
					xtick=data,
					hide axis,
					x tick label style={rotate=60,anchor=east},
					ybar,
					bar width=7.5pt,
					ymin=0,
					ymax=100,
					width=\textwidth, 
					enlarge x limits={abs=0.5*\pgfplotbarwidth},					
					height=9cm, width=16cm,
					]
					\addplot [fill=blue] coordinates {
						(AP Calculus BC,1)
						(AMC 12,4)
						(Codeforces Rating,5)
						(AP English Literature,8)
						(AMC 10,10)
						(Uniform Bar Exam,10)
						(AP English Language,14)
						(AP Chemistry,22)
						(GRE Quantitative,25)
						(AP Physics 2,30)
						(USABO Semifinal 2020,31)
						(AP Macroeconomics,33)
						(LSAT,40)
						(AP Statistics,40)
						(VNHSGE Chemistry,0)
						(GRE Writing,54)
						(VNHSGE History,0)
						(VNHSGE Biology,0)
						(VNHSGE Mathematics,0)
						(AP Microeconomics,60)
						(AP Biology,62)
						(VNHSGE Physics,0)
						(VNHSGE Geography,0)
						(GRE Verbal,63)
						(AP World History,65)
						(VNHSGE Literature,0)
						(SAT Math,70)
						(VNHSGE Civic Education,0)
						(AP US History,74)
						(AP US Government,77)
						(VNHSGE English,0)
						(AP Psychology,85)
						(AP Art History,86)
						(SAT EBRW,86)
						(AP Environmental Science,90)	
					};					
				\end{axis}
				\begin{axis}[
					ylabel={Percentage},
					symbolic x coords={
						AP Calculus BC,
						AMC 12,
						Codeforces Rating,
						AP English Literature,
						AMC 10,
						Uniform Bar Exam,
						AP English Language,
						AP Chemistry,
						GRE Quantitative,
						AP Physics 2,
						USABO Semifinal 2020,
						AP Macroeconomics,
						LSAT,
						AP Statistics,
						VNHSGE Chemistry,
						GRE Writing,
						VNHSGE History,
						VNHSGE Biology,
						VNHSGE Mathematics,
						AP Microeconomics,
						AP Biology,
						VNHSGE Physics,
						VNHSGE Geography,
						GRE Verbal,
						AP World History,
						VNHSGE Literature,
						SAT Math,
						VNHSGE Civic Education,
						AP US History,
						AP US Government,
						VNHSGE English,
						AP Psychology,
						AP Art History,
						SAT EBRW,
						AP Environmental Science,
					},
					xtick=data,
					x tick label style={rotate=60,anchor=east},
					hide axis,
					ybar,
					bar width=7.5pt,
					ymin=0,
					ymax=100,
					width=\textwidth, 
					enlarge x limits={abs=0.5*\pgfplotbarwidth},					
					height=9cm, width=16cm,
					]	
					\addplot [fill=red] coordinates {
						(AP Calculus BC,0)
						(AMC 12,0)
						(Codeforces Rating,0)
						(AP English Literature,0)
						(AMC 10,0)
						(Uniform Bar Exam,0)
						(AP English Language,0)
						(AP Chemistry,0)
						(GRE Quantitative,0)
						(AP Physics 2,0)
						(USABO Semifinal 2020,0)
						(AP Macroeconomics,0)
						(LSAT,0)
						(AP Statistics,00)
						(VNHSGE Chemistry,52.5)
						(GRE Writing,0)
						(VNHSGE History,88.5)
						(VNHSGE Biology,69)
						(VNHSGE Mathematics,60)
						(AP Microeconomics,0)
						(AP Biology,0)
						(VNHSGE Physics,66)
						(VNHSGE Geography,79.5)
						(GRE Verbal,0)
						(AP World History,0)
						(VNHSGE Literature,56.8)
						(SAT Math,0)
						(VNHSGE Civic Education,88.5)
						(AP US History,0)
						(AP US Government,0)
						(VNHSGE English,92)
						(AP Psychology,0)
						(AP Art History,0)
						(SAT EBRW,0)
						(AP Environmental Science,0)	
					};
					
				\end{axis}
				\begin{axis}[
					ylabel={Percentage},
					symbolic x coords={
						AP Calculus BC,
						AMC 12,
						Codeforces Rating,
						AP English Literature,
						AMC 10,
						Uniform Bar Exam,
						AP English Language,
						AP Chemistry,
						GRE Quantitative,
						AP Physics 2,
						USABO Semifinal 2020,
						AP Macroeconomics,
						LSAT,
						AP Statistics,
						VNHSGE Chemistry,
						GRE Writing,
						VNHSGE History,
						VNHSGE Biology,
						VNHSGE Mathematics,
						AP Microeconomics,
						AP Biology,
						VNHSGE Physics,
						VNHSGE Geography,
						GRE Verbal,
						AP World History,
						VNHSGE Literature,
						SAT Math,
						VNHSGE Civic Education,
						AP US History,
						AP US Government,
						VNHSGE English,
						AP Psychology,
						AP Art History,
						SAT EBRW,
						AP Environmental Science,
					},
					x tick label style={rotate=60,anchor=east},
					hide axis,
					ybar,
					bar width=7.5pt,
					ymin=0,
					ymax=100,
					width=\textwidth, 
					enlarge x limits={abs=0.5*\pgfplotbarwidth},					
					height=9cm, width=16cm,
					]	
					\addplot [fill=yellow] coordinates {
						(AP Calculus BC,0)
						(AMC 12,0)
						(Codeforces Rating,0)
						(AP English Literature,0)
						(AMC 10,0)
						(Uniform Bar Exam,0)
						(AP English Language,0)
						(AP Chemistry,0)
						(GRE Quantitative,0)
						(AP Physics 2,0)
						(USABO Semifinal 2020,0)
						(AP Macroeconomics,0)
						(LSAT,0)
						(AP Statistics,00)
						(VNHSGE Chemistry,48)
						(GRE Writing,0)
						(VNHSGE History,56.5)
						(VNHSGE Biology,58)
						(VNHSGE Mathematics,58.8)
						(AP Microeconomics,0)
						(AP Biology,0)
						(VNHSGE Physics,61)
						(VNHSGE Geography,61.5)
						(GRE Verbal,0)
						(AP World History,0)
						(VNHSGE Literature,67.96)
						(SAT Math,0)
						(VNHSGE Civic Education,70.5)
						(AP US History,0)
						(AP US Government,0)
						(VNHSGE English,79)
						(AP Psychology,0)
						(AP Art History,0)
						(SAT EBRW,0)
						(AP Environmental Science,0)	
					};
				\end{axis}
			\end{tikzpicture}
		\end{center}
		\caption{Performance of ChatGPT, BingChat on VNHSGE dataset and GPT-3.5, GPT-4 on other datasets (Figure from~\cite{dao2023vnhsge}).}
		\label{fig:chatgpt_exam}
	\end{figure*}
	
	\section{Objectives and Methods}
	
	The objective of this study is to assess the efficacy of LLMs in successfully completing the Vietnamese National High School Graduation Examination (VNHSGE) and offer a comprehensive analysis of the obtained outcomes. In this study, we conduct a comparative analysis of the performance exhibited by ChatGPT and BingChat. Additionally, we examined the performance of LLMs in relation to Vietnamese students' performance on the VNHSGE exanimation. This study assessed the performance of LLMs utilizing an evaluation set derived from the VNHSGE dataset~\cite{dao2023vnhsge}. The evaluation set consists of questions from the VNHSGE examination spanning the years 2019 to 2023, covering a range of subjects including mathematics, English, physics, chemistry, biology, history, geography, civic education, and literature. The evaluation set consists of a total of 30 essay questions specifically designed for the literature subject, while the remaining 1700 questions are in the form of multiple-choice for other subjects. The primary objective of this study is to make a valuable contribution to the advancement of AI tools designed to provide educational support. Additionally, this research seeks to provide insights into the potential future applications of AI in reshaping the educational environment.

	\section{Results}

	\subsection{Performance} 
	
	The findings of the investigation indicated that BingChat exhibits superior performance compared to ChatGPT across all subjects, with the exception of literature. While there is no discernible disparity in their academic performance in the fields of mathematics, physics, and chemistry, BingChat demonstrates exceptional proficiency in various other subjects.
	
	Table~\ref{tabl:llm_result} presents the performance metrics of ChatGPT and BingChat across multiple domains. The performance in subjects that necessitate intricate computation and reasoning, such as mathematics, physics, chemistry, and biology, exhibits a range spanning from $48\%$ to $69\%$. The academic disciplines that place significant emphasis on language proficiency, such as literature, English, history, geography, and civic education, exhibit a range of achievement levels spanning from $56.5\%$ to $92.4\%$. Language models such as ChatGPT and BingChat have undergone extensive training on copious quantities of textual data sourced from diverse domains. Nevertheless, it is possible that they may have a deficiency in subject-specific knowledge. Disciplines such as mathematics, physics, chemistry, and biology frequently demand a profound understanding and sophisticated computational skills that may surpass the capabilities of language models. Conversely, disciplines such as literature, English, history, geography, and civic education frequently demand proficient language skills and the capacity to comprehend intricate texts, domains in which language models may possess adequate capabilities

	According to the data presented in Figure~\ref{fig:ChatGPT_BingChat_exam}, it can be observed that BingChat exhibits superior performance compared to ChatGPT across various subjects, with the exception of literature. There exists a limited disparity between the two in disciplines that necessitate substantial computation and logical thinking, such as mathematics, physics, and chemistry. Nevertheless, ChatGPT outperforms BingChat in the domain of literature, potentially due to the fact that BingChat primarily functions as a search engine, thereby potentially rendering its outcomes less tailored to the specific requirements of literary inquiries. In the remaining subjects, BingChat demonstrates superior performance compared to ChatGPT. It is noteworthy to mention that BingChat utilizes the GPT-4 model, whereas ChatGPT is built upon GPT-3.5. BingChat demonstrates potential in providing precise responses by leveraging publicly accessible online information.
	
	Figure~\ref{fig:chatgpt_exam} contrasts the performance of ChatGPT and BingChat on the VNHSGE dataset to other datasets included in the GPT-4 Report~\cite{OpenAI_gpt_4_report}. ChatGPT exhibits comparable performance to GPT-3.5 across diverse domains, whereas BingChat demonstrates enhanced proficiency specifically in text-oriented subjects. Nevertheless, it is worth noting that BingChat does not exhibit a substantial superiority over ChatGPT in domains that necessitate intricate computational abilities and logical reasoning. The performance of GPT-4 surpasses that of GPT-3.5 in comparable tasks, potentially attributable to the question structure within the VNHSGE dataset.
	
	\subsection{Human-like Level}

	\begin{figure*}[ht!]
		\begin{center}
			\begin{tikzpicture}  
				\begin{axis}  
					[  
					ybar, 
					bar width=15pt,
					enlargelimits=0.15, 
					legend style={at={(0.5,-0.175)}, 	
						anchor=north,legend columns=-1},       
					ylabel={Mathematics Score}, 
					symbolic x coords={
						2019,
						2020,
						2021,
						2022,
						2023,
					},  
					xtick=data,  
					nodes near coords,  
					every node near coord/.append style={rotate=90, anchor=west},  
					width=\textwidth, 
					height=3.75cm, 
					width=16cm,
					axis x line*=bottom, axis y line*=left
					]  
					\addplot [fill=yellow] coordinates {
						(2019,5.2)
						(2020,6.6)
						(2021,6)
						(2022,6.2)
						(2023,5.4)
					};
					\addplot [fill=red] coordinates {
						(2019,5.6)
						(2020,5.6)
						(2021,6.6)
						(2022,6)
						(2023,6.2)						
					};
					\addplot [fill=green] coordinates {
						(2019,5.64)
						(2020,6.67)
						(2021,6.61)
						(2022,6.47)
						(2023,6.25)		
					};
					\addplot [fill=blue] coordinates {
						(2019,6.4)
						(2020,7.8)
						(2021,7.8)
						(2022,7.8)
						(2023,7.6)
						
					};
				\end{axis}  
			\end{tikzpicture}
			\begin{tikzpicture}  
				\begin{axis}  
					[  
					ybar, 
					bar width=15pt,
					enlargelimits=0.15, 
					legend style={at={(0.5,-0.175)}, 	
						anchor=north,legend columns=-1},       
					ylabel={Literature Score}, 
					symbolic x coords={
						2019,
						2020,
						2021,
						2022,
						2023,
					},  
					xtick=data,  
					nodes near coords,  
					every node near coord/.append style={rotate=90, anchor=west},  
					width=\textwidth, 
					height=3.75cm, 
					width=16cm,
					axis x line*=bottom, axis y line*=left
					]  
					\addplot [fill=yellow] coordinates {
						(2019,7.5)
						(2020,6.89)
						(2021,7.5)
						(2022,5.63)
						(2023,6.48)
						
					};
					\addplot [fill=red] coordinates {
						(2019,5.28)
						(2020,5.13)
						(2021,6.03)
						(2022,7)
						(2023,5)
						
					};
					\addplot [fill=green] coordinates {
						(2019,5.49)
						(2020,6.61)
						(2021,6.47)
						(2022,6.51)
						(2023,6.86)
						
					};
					\addplot [fill=blue] coordinates {
						(2019,6)
						(2020,7)
						(2021,7)
						(2022,7)
						(2023,7)
						
					};
				\end{axis}  
			\end{tikzpicture}
			\begin{tikzpicture}  
				\begin{axis}  
					[  
					ybar, 
					bar width=15pt,
					enlargelimits=0.15, 
					legend style={at={(0.5,-0.25)}, 	
						anchor=north,legend columns=-1},       
					ylabel={English Score}, 
					symbolic x coords={
						2019,
						2020,
						2021,
						2022,
						2023,
					},  
					xtick=data,  
					nodes near coords,  
					every node near coord/.append style={rotate=90, anchor=west},  
					width=\textwidth, 
					height=3.75cm, 
					width=16cm,
					axis x line*=bottom, axis y line*=left
					]  
					\addplot [fill=yellow] coordinates {
						(2019,7.6)
						(2020,8.6)
						(2021,7.6)
						(2022,8)
						(2023,7.8)
					};
					\addplot [fill=red] coordinates {
						(2019,9.2)
						(2020,9.6)
						(2021,8.6)
						(2022,9.4)
						(2023,9.4)						
					};
					\addplot [fill=green] coordinates {
						(2019,4.36)
						(2020,4.58)
						(2021,5.84)
						(2022,5.15)
						(2023,5.45)							
					};
					\addplot [fill=blue] coordinates {
						(2019,3.2)
						(2020,3.4)
						(2021,4)
						(2022,3.8)
						(2023,4.2)
					};
					\legend{ChatGPT, BingChat, AVS, MVS} 
				\end{axis}  
			\end{tikzpicture} 
		\end{center}
		\caption{Comparison in core subjects.} 
		\label{fig:core subjects}
	\end{figure*}
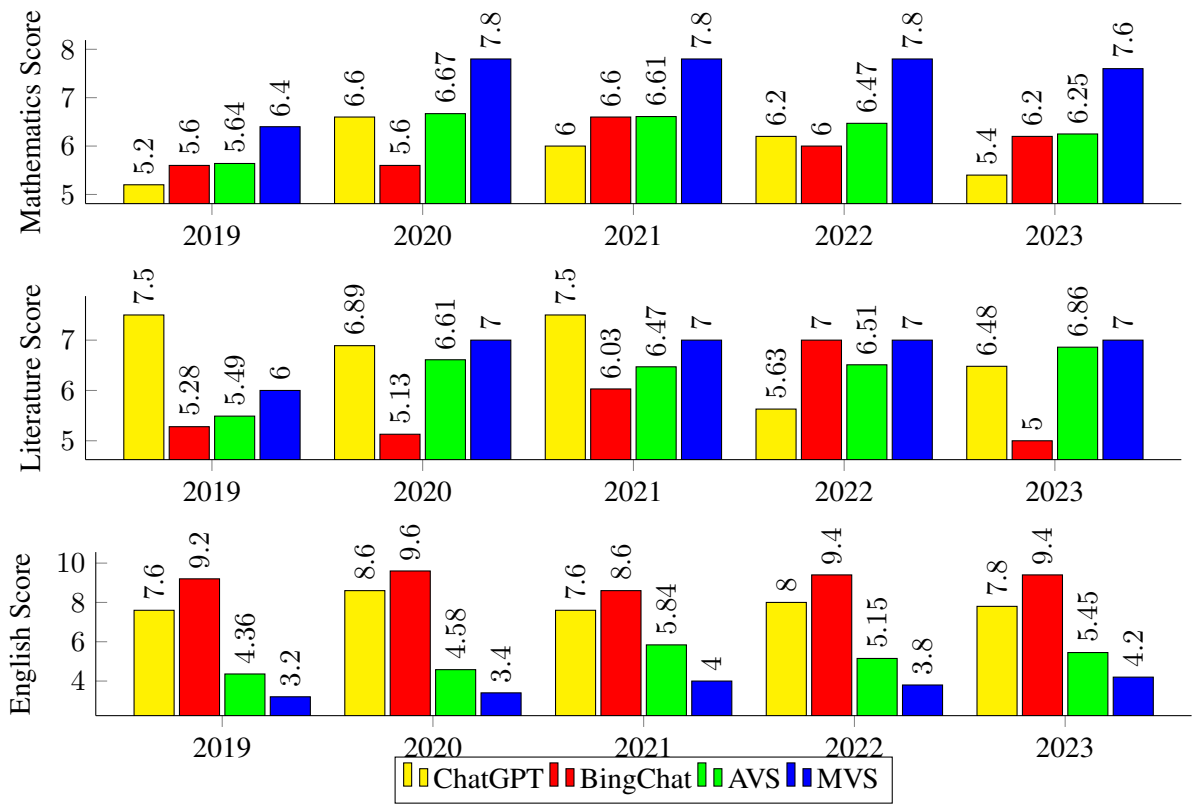

	Next, we assess the comparative abilities of LLMs and humans, despite the inherent challenges associated with making such a comparison due to the distinct contextual factors involved. This comparative analysis can facilitate the evaluation of the potential efficacy of LLMs as educational tools for Vietnamese students across diverse academic disciplines.
	
	Figure~\ref{fig:core subjects} depicts a comparative analysis of the academic performance across three entities: ChatGPT, BingChat, and Vietnamese students. The evaluation focuses on three fundamental disciplines, namely mathematics, literature, and English. These subjects are essential components of the examination and are mandatory for all students. The examination of LLMs and Vietnamese students highlights divergent levels of academic achievement in distinct disciplines. In the field of mathematics, both ChatGPT and BingChat encounter difficulties when faced with intricate problems, ultimately demonstrating limitations in comparison to human students. Nevertheless, within the realm of high school literature, both ChatGPT and BingChat exhibit a level of performance that is comparable to that of a human. However, it is important to note that their abilities in producing emotionally nuanced essays and conducting comprehensive literary analyses are somewhat restricted. In the realm of English language learning, ChatGPT and BingChat exhibit superior performance compared to Vietnamese students, thereby offering effective support for language acquisition.
	
	\textbf{Mathematics}: According to the findings, ChatGPT and BingChat are unable to match the performance of human students in Vietnam's high school mathematics curriculum. Although they possess extensive training on textual data derived from the internet, they encounter difficulties when confronted with intricate mathematical problems, albeit demonstrate proficiency in handling more elementary mathematical concepts~\cite{dao2023math}. The mathematics questions at the high school level necessitate the utilization of reasoning, logical thinking, analytical aptitude, and the capacity to apply acquired knowledge in real-world scenarios. In order to attain a level of mathematical proficiency comparable to that of humans in the context of high school education, both ChatGPT and BingChat necessitate significant enhancements.
	
	\textbf{Literature}: Both ChatGPT and BingChat have undergone comprehensive training using extensive datasets in the Vietnamese language, which has equipped them with the ability to proficiently analyze and generate essays. When considering high school literature, the proficiency exhibited by LLMs, including ChatGPT and BingChat, can be described as comparable to that of human performance. However, it is important to highlight that ChatGPT and BingChat lack the capability to generate emotionally nuanced essays or engage in comprehensive literary analyses~\cite{dao2023vnhsge}. In essence, ChatGPT can be regarded as a resource that aids Vietnamese students in their pursuit of literary studies.
	
	\textbf{English}: Based on the findings, it was observed that ChatGPT and BingChat exhibited superior performance in the domain of high school English when compared to Vietnamese students~\cite{dao2023eng,dao2023performance}. It is noteworthy to acknowledge that the English proficiency of Vietnamese students is comparatively lower than the global average. ChatGPT and BingChat have proven to be efficacious resources that Vietnamese students can employ for the purpose of acquiring proficiency in foreign languages.
	
\begin{figure*}[ht!]
		\begin{center}
			\begin{tikzpicture}  
				\begin{axis}  
					[  
					ybar, 
					bar width=15pt,
					enlargelimits=0.15, 
					legend style={at={(0.5,-0.25)}, 	
						anchor=north,legend columns=-1},          
					ylabel={Physics Score}, 
					symbolic x coords={
						2019,
						2020,
						2021,
						2022,
						2023,
					},  
					xtick=data,  
					nodes near coords,  
					every node near coord/.append style={rotate=90, anchor=west},  
					width=\textwidth, 
					height=3.75cm, 
					width=16cm,
					axis x line*=bottom, axis y line*=left
					]  
					
					\addplot [fill=yellow] coordinates {
						(2019,6.0)
						(2020,6.25)
						(2021,6.0)
						(2022,6.5)
						(2023,5.75)
					};
					\addplot [fill=red] coordinates {
						(2019,5.5)
						(2020,6.75)
						(2021,6.75)
						(2022,6.75)
						(2023,7.25)
						
					};
					\addplot [fill=green] coordinates {
						(2019,5.57)
						(2020,6.72)
						(2021,6.56)
						(2022,6.72)
						(2023,6.57)
						
					};
					\addplot [fill=blue] coordinates {
						(2019,6.25)
						(2020,7.75)
						(2021,7.5)
						(2022,7.25)
						(2023,7.5)
						
					};
					
				\end{axis}  
			\end{tikzpicture} 
			\begin{tikzpicture}  
				\begin{axis}  
					[  
					ybar, 
					bar width=15pt,
					enlargelimits=0.15, 
					legend style={at={(0.5,-0.175)}, 	
						anchor=north,legend columns=-1},       
					ylabel={Chemistry Score}, 
					symbolic x coords={
						2019,
						2020,
						2021,
						2022,
						2023,
					},  
					xtick=data,  
					nodes near coords,  
					every node near coord/.append style={rotate=90, anchor=west},  
					width=\textwidth, 
					height=3.75cm, 
					width=16cm,
					axis x line*=bottom, axis y line*=left
					]  
					\addplot [fill=yellow] coordinates {
						(2019,4)
						(2020,4.25)
						(2021,6.25)
						(2022,4.75)
						(2023,4.75)
					};
					\addplot [fill=red] coordinates {
						(2019,5.5)
						(2020,5.75)
						(2021,5)
						(2022,4.75)
						(2023,5.25)						
					};
					\addplot [fill=green] coordinates {
						(2019,5.35)
						(2020,6.71)
						(2021,6.63)
						(2022,6.7)
						(2023,6.74)
						
					};
					\addplot [fill=blue] coordinates {
						(2019,6)
						(2020,7.75)
						(2021,7.75)
						(2022,8)
						(2023,7.5)

					};
				\end{axis}  
			\end{tikzpicture} 
			\begin{tikzpicture}  
				\begin{axis}  
					[  
					ybar, 
					bar width=15pt,
					enlargelimits=0.15, 
					legend style={at={(0.5,-0.25)}, 	
						anchor=north,legend columns=-1},       
					ylabel={Biology Score}, 
					symbolic x coords={
						2019,
						2020,
						2021,
						2022,
						2023,
					},  
					xtick=data,  
					nodes near coords,  
					every node near coord/.append style={rotate=90, anchor=west},  
					width=\textwidth, 
					height=3.75cm, 
					width=16cm,
					axis x line*=bottom, axis y line*=left
					]  
					\addplot [fill=yellow] coordinates {
						(2019,6)
						(2020,6)
						(2021,5.25)
						(2022,5.75)
						(2023,6)
					};
					\addplot [fill=red] coordinates {
						(2019,6.75)
						(2020,7.25)
						(2021,6.75)
						(2022,7.25)
						(2023,6.5)						
					};
					\addplot [fill=green] coordinates {
						(2019,4.68)
						(2020,5.6)
						(2021,5.51)
						(2022,5.02)
						(2023,6.39)					
					};
					\addplot [fill=blue] coordinates {
						(2019,4.5)
						(2020,5.25)
						(2021,5.25)
						(2022,4.5)
						(2023,6.5)
					};
					\legend{ChatGPT, BingChat, AVS, MVS} 
				\end{axis}  
			\end{tikzpicture} 
		\end{center}
		\caption{Comparison in natural combination.} 
		\label{fig:natural combination}
	\end{figure*}
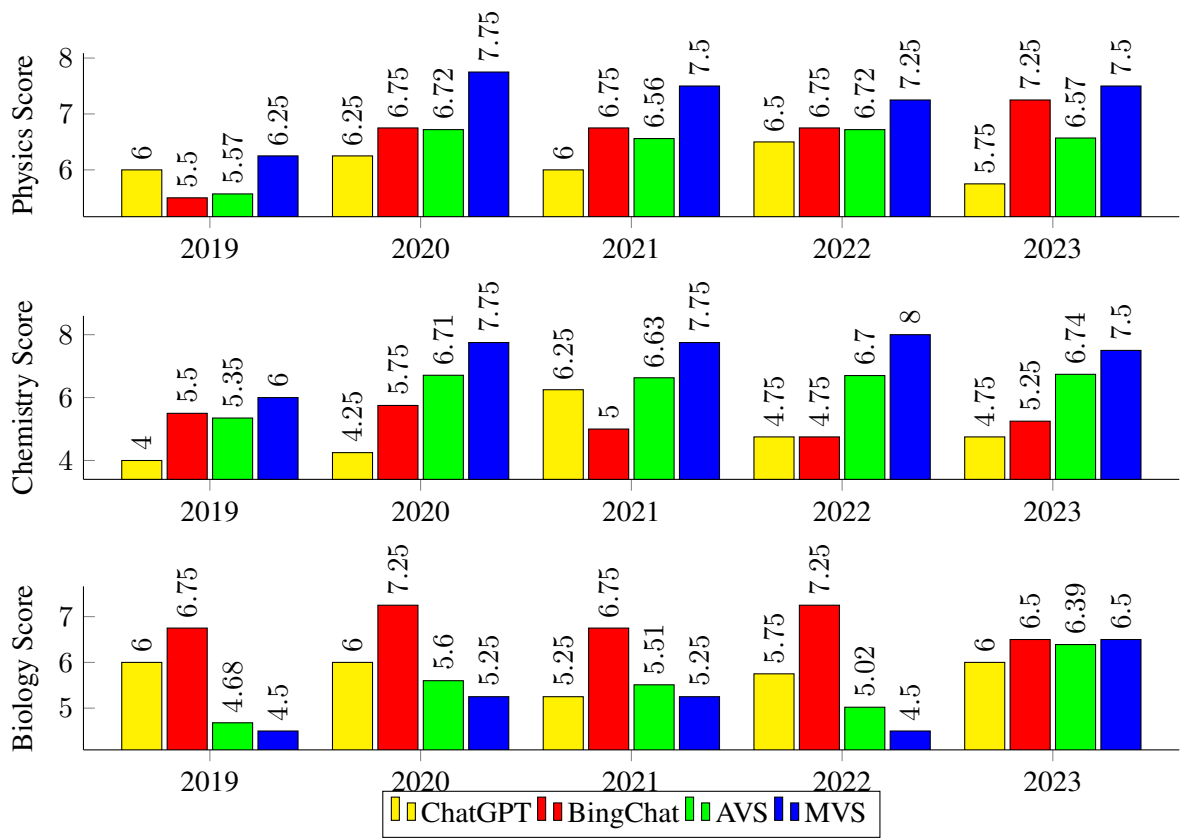

	Figure~\ref{fig:natural combination} depicts a comparison of the performance among ChatGPT, BingChat, and Vietnamese students in the natural combination, including physics, chemistry, and biology. In physics, ChatGPT and BingChat perform comparably to the average score of Vietnamese students. However, they still lag behind top-performing students, particularly in intricate computations and reasoning. For chemistry, these language models offer relevant knowledge and solutions, but they lack the expertise for complex problem-solving. In biology, ChatGPT and BingChat outperform Vietnamese students, serving as valuable resources for understanding fundamental concepts.
	
	\textbf{Physics}: The performance of ChatGPT and BingChat exhibits a level of similarity to the mean score achieved by Vietnamese students in the field of physics~\cite{dao2023vnhsge,dao2023phy}. Nevertheless, the scores attained by most Vietnamese students surpass those achieved by the aforementioned LLMs. LLMs, who possess comprehensive training in the discipline of physics, are capable of offering precise responses and profound elucidations to aid students in comprehending the subject matter of physics. The existing models, nevertheless, necessitate further refinement, particularly in relation to physics matters that necessitate complex calculations and logical reasoning.
	
	\textbf{Chemistry}: As shown in Figure~\ref{fig:natural combination}, ChatGPT and BingChat do not exhibit an equivalent level of expertise in this subject matter~\cite{dao2023vnhsge,dao2023che}. Although these LLMs may offer valuable knowledge and solutions in the realm of chemistry, their proficiency may be insufficient for resolving intricate chemistry problems that necessitate advanced levels of analysis and reasoning. Nevertheless, when it comes to imparting theoretical knowledge and information, LLMs have the potential to serve as valuable resources for Vietnamese high school students studying chemistry.
	
	\textbf{Biology}: The results suggest that both ChatGPT and BingChat demonstrate superior performance compared to Vietnamese students in the field of biology~\cite{dao2023vnhsge}. It is noteworthy to acknowledge that biology holds a lower priority among many Vietnamese students in comparison to subjects such as mathematics, physics, and chemistry. Vietnamese students exhibit lower performance in mathematics, physics, and chemistry compared to their biology scores. LLMs possess the capacity to tackle fundamental inquiries within the realm of biology, encompassing the elucidation of definitions, comprehension of concepts, rudimentary problem-solving, and the provision of specific instances. Consequently, the utilization of LLMs can prove to be advantageous in aiding high school students in their understanding of fundamental concepts and challenges in the field of biology.
	
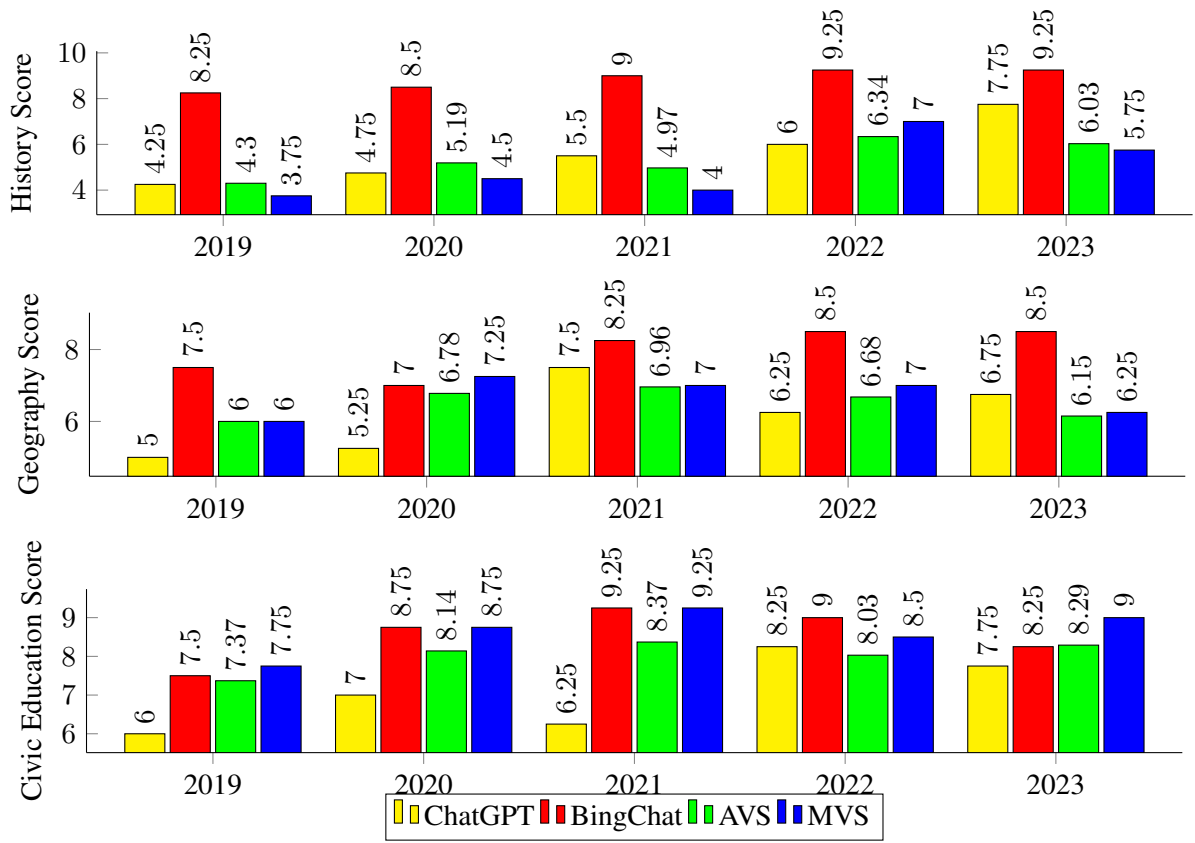
\begin{figure*}[ht!]
		\begin{center}
			\begin{tikzpicture}  
				\begin{axis}  
					[  
					ybar, 
					bar width=15pt,
					enlargelimits=0.15, 
					legend style={at={(0.5,-0.175)}, 	
						anchor=north,legend columns=-1},       
					ylabel={History Score}, 
					symbolic x coords={
						2019,
						2020,
						2021,
						2022,
						2023,
					},  
					xtick=data,  
					nodes near coords,  
					every node near coord/.append style={rotate=90, anchor=west},  
					width=\textwidth, 
					height=3.75cm, 
					width=16cm,
					axis x line*=bottom, axis y line*=left
					]  
					\addplot [fill=yellow] coordinates {
						(2019,4.25)
						(2020,4.75)
						(2021,5.5)
						(2022,6)
						(2023,7.75)	
					};
					\addplot [fill=red] coordinates {
						(2019,8.25)
						(2020,8.5)
						(2021,9)
						(2022,9.25)
						(2023,9.25)						
					};
					\addplot [fill=green] coordinates {
						(2019,4.3)
						(2020,5.19)
						(2021,4.97)
						(2022,6.34)
						(2023,6.03)		
					};
					\addplot [fill=blue] coordinates {
						(2019,3.75)
						(2020,4.5)
						(2021,4)
						(2022,7)
						(2023,5.75)
					};
				\end{axis}  
			\end{tikzpicture}
			\begin{tikzpicture}  
				\begin{axis}  
					[  
					ybar, 
					bar width=15pt,
					enlargelimits=0.15, 
					legend style={at={(0.5,-0.175)}, 	
						anchor=north,legend columns=-1},       
					ylabel={Geography Score}, 
					symbolic x coords={
						2019,
						2020,
						2021,
						2022,
						2023,
					},  
					xtick=data,  
					nodes near coords,  
					every node near coord/.append style={rotate=90, anchor=west},  
					width=\textwidth, 
					height=3.75cm, 
					width=16cm,
					axis x line*=bottom, axis y line*=left
					]  
					\addplot [fill=yellow] coordinates {
						(2019,5)
						(2020,5.25)
						(2021,7.5)
						(2022,6.25)
						(2023,6.75)
						
					};
					\addplot [fill=red] coordinates {
						(2019,7.5)
						(2020,7)
						(2021,8.25)
						(2022,8.5)
						(2023,8.5)
						
					};
					\addplot [fill=green] coordinates {
						(2019,6)
						(2020,6.78)
						(2021,6.96)
						(2022,6.68)
						(2023,6.15)
						
					};
					\addplot [fill=blue] coordinates {
						(2019,6)
						(2020,7.25)
						(2021,7)
						(2022,7)
						(2023,6.25)

					};
				\end{axis}  
			\end{tikzpicture} 
			\begin{tikzpicture}  
				\begin{axis}  
					[  
					ybar, 
					bar width=15pt,
					enlargelimits=0.15, 
					legend style={at={(0.5,-0.25)}, 	
						anchor=north,legend columns=-1},       
					ylabel={Civic Education Score}, 
					symbolic x coords={
						2019,
						2020,
						2021,
						2022,
						2023,
					},  
					xtick=data,  
					nodes near coords,  
					every node near coord/.append style={rotate=90, anchor=west},  
					width=\textwidth, 
					height=3.75cm, 
					width=16cm,
					axis x line*=bottom, axis y line*=left
					]  
					\addplot [fill=yellow] coordinates {
						(2019,6)
						(2020,7)
						(2021,6.25)
						(2022,8.25)
						(2023,7.75)
						
					};
					\addplot [fill=red] coordinates {
						(2019,7.5)
						(2020,8.75)
						(2021,9.25)
						(2022,9)
						(2023,8.25)
						
					};
					\addplot [fill=green] coordinates {
						(2019,7.37)
						(2020,8.14)
						(2021,8.37)
						(2022,8.03)
						(2023,8.29)
						
					};
					\addplot [fill=blue] coordinates {
						(2019,7.75)
						(2020,8.75)
						(2021,9.25)
						(2022,8.5)
						(2023,9)	
					};
					\legend{ChatGPT, BingChat, AVS, MVS} 
				\end{axis}  
			\end{tikzpicture} 
		\end{center}
		\caption{Comparison in social combination.} 
		\label{fig:social combination}
	\end{figure*}

	Figure~\ref{fig:social combination} presents a comparison of the performance among ChatGPT, BingChat, and Vietnamese students in the social combination, including history, geography, and civic education. Both ChatGPT and BingChat demonstrate efficacy in providing informative responses in the field of history, with BingChat exhibiting a slightly superior performance. In the field of geography, both models exhibit comparable performance in facilitating the learning of concepts and terminology among Vietnamese students. In the realm of civic education, both ChatGPT and BingChat demonstrate remarkable capacities akin to human intelligence, thereby offering substantial insights into the domains of politics, law, civil liberties, and societal concerns.

	\textbf{History}: The performance of BingChat is superior, while ChatGPT's outcomes are on par with those achieved by Vietnamese students~\cite{dao2023vnhsge}. ChatGPT and BingChat possess the capability to comprehend and analyze a wide range of historical inquiries, thanks to their extensive and varied training datasets. Consequently, they are capable of generating coherent and valuable replies in response to such queries. While ChatGPT and BingChat may face certain difficulties when confronted with intricate inquiries, they possess the potential to serve as valuable educational tools for high school students studying history.
	
	\textbf{Geography}: BingChat demonstrates superior performance, while ChatGPT exhibits comparable performance to Vietnamese students~\cite{dao2023vnhsge}. The findings suggest that both ChatGPT and BingChat possess the ability to comprehend and provide appropriate responses to geography inquiries at the high school level. The utilization of these instructional methods can significantly enhance students' comprehension and retention of geography concepts and terminology, thereby fostering a more effective learning experience within the context of high school geography education. Nevertheless, these individuals might encounter constraints when confronted with intricate and profound inquiries that necessitate sophisticated critical analysis.
	
	\textbf{Civic Education}: BingChat and ChatGPT demonstrate human-like capabilities within the domain of civic education~\cite{dao2023vnhsge}. Due to their extensive background in civic education and law-related disciplines, these LLMs possess the necessary expertise to impart knowledge at the high school level in various domains, including politics, law, citizen rights and responsibilities, and other pertinent social matters. Hence, ChatGPT and BingChat can serve as valuable reference tools for Vietnamese students participating in the study of civic education. 
	
	\subsection{Score}
	
	\begingroup
	\renewcommand{\arraystretch}{1.5} 
	
	\begin{table*}[ht!]
		\caption{Subject scores and combination scores for natural and social combinations}
		\label{tab:GAS}
		\centering
		\resizebox{\textwidth}{!}{%
			\begin{tabular}{l|ccccccc|ccccccc|}
				\cline{2-15}
				& \multicolumn{7}{c|}{\textbf{ChatGPT}~\cite{dao2023exa}}                                                                                                                                                                                                & \multicolumn{7}{c|}{\textbf{BingChat}}                                                                                                                                                                                               \\ \cline{2-15} 
				& \multicolumn{1}{c|}{\textbf{M}} & \multicolumn{1}{c|}{\textbf{L}} & \multicolumn{1}{c|}{\textbf{E}} & \multicolumn{1}{c|}{\textbf{$\mathrm{AC_N}$}} & \multicolumn{1}{c|}{\textbf{$\mathrm{AC_S}$}} & \multicolumn{1}{c|}{\textbf{$\mathrm{GAS_N}$}} & \textbf{$\mathrm{GAS_S}$} & \multicolumn{1}{c|}{\textbf{M}} & \multicolumn{1}{c|}{\textbf{L}} & \multicolumn{1}{c|}{\textbf{E}} & \multicolumn{1}{c|}{\textbf{$\mathrm{AC_N}$}} & \multicolumn{1}{c|}{\textbf{$\mathrm{AC_S}$}} & \multicolumn{1}{c|}{\textbf{$\mathrm{GAS_N}$}} & \textbf{$\mathrm{GAS_S}$} \\ \hline
				\multicolumn{1}{|l|}{\textbf{2019}} & \multicolumn{1}{c|}{5.2}        & \multicolumn{1}{c|}{7.5}        & \multicolumn{1}{c|}{7.6}        & \multicolumn{1}{c|}{5.33}          & \multicolumn{1}{c|}{5.08}          & \multicolumn{1}{c|}{6.41}           & 6.35           & \multicolumn{1}{c|}{5.6}        & \multicolumn{1}{c|}{5.28}      & \multicolumn{1}{c|}{9.2}        & \multicolumn{1}{c|}{5.92}          & \multicolumn{1}{c|}{7.75}          & \multicolumn{1}{c|}{6.50}           & 6.96           \\ \hline
				\multicolumn{1}{|l|}{\textbf{2020}} & \multicolumn{1}{c|}{6.6}        & \multicolumn{1}{c|}{6.89}       & \multicolumn{1}{c|}{8.6}        & \multicolumn{1}{c|}{5.50}          & \multicolumn{1}{c|}{5.67}          & \multicolumn{1}{c|}{6.89}           & 6.94           & \multicolumn{1}{c|}{5.6}        & \multicolumn{1}{c|}{5.123}      & \multicolumn{1}{c|}{9.6}        & \multicolumn{1}{c|}{6.58}          & \multicolumn{1}{c|}{8.08}          & \multicolumn{1}{c|}{6.73}           & 7.10           \\ \hline
				\multicolumn{1}{|l|}{\textbf{2021}} & \multicolumn{1}{c|}{6}          & \multicolumn{1}{c|}{7.5}        & \multicolumn{1}{c|}{7.6}        & \multicolumn{1}{c|}{5.83}          & \multicolumn{1}{c|}{6.42}          & \multicolumn{1}{c|}{6.73}           & 6.88           & \multicolumn{1}{c|}{6.6}        & \multicolumn{1}{c|}{6.03}      & \multicolumn{1}{c|}{8.6}        & \multicolumn{1}{c|}{6.17}          & \multicolumn{1}{c|}{8.83}          & \multicolumn{1}{c|}{6.85}           & 7.51           \\ \hline
				\multicolumn{1}{|l|}{\textbf{2022}} & \multicolumn{1}{c|}{6.2}        & \multicolumn{1}{c|}{5.63}       & \multicolumn{1}{c|}{8}          & \multicolumn{1}{c|}{5.67}          & \multicolumn{1}{c|}{6.83}          & \multicolumn{1}{c|}{6.37}           & 6.67           & \multicolumn{1}{c|}{6}          & \multicolumn{1}{c|}{7}          & \multicolumn{1}{c|}{9.4}        & \multicolumn{1}{c|}{6.25}          & \multicolumn{1}{c|}{8.92}          & \multicolumn{1}{c|}{7.16}           & 7.83           \\ \hline
				\multicolumn{1}{|l|}{\textbf{2023}} & \multicolumn{1}{c|}{5.4}        & \multicolumn{1}{c|}{6.48}       & \multicolumn{1}{c|}{7.8}        & \multicolumn{1}{c|}{5.50}          & \multicolumn{1}{c|}{7.42}          & \multicolumn{1}{c|}{6.30}           & 6.77           & \multicolumn{1}{c|}{6.2}        & \multicolumn{1}{c|}{4.98}      & \multicolumn{1}{c|}{9.4}        & \multicolumn{1}{c|}{6.33}          & \multicolumn{1}{c|}{8.67}          & \multicolumn{1}{c|}{6.73}           & 7.31           \\ \hline
			\end{tabular}
		}
	\end{table*}
	\endgroup

	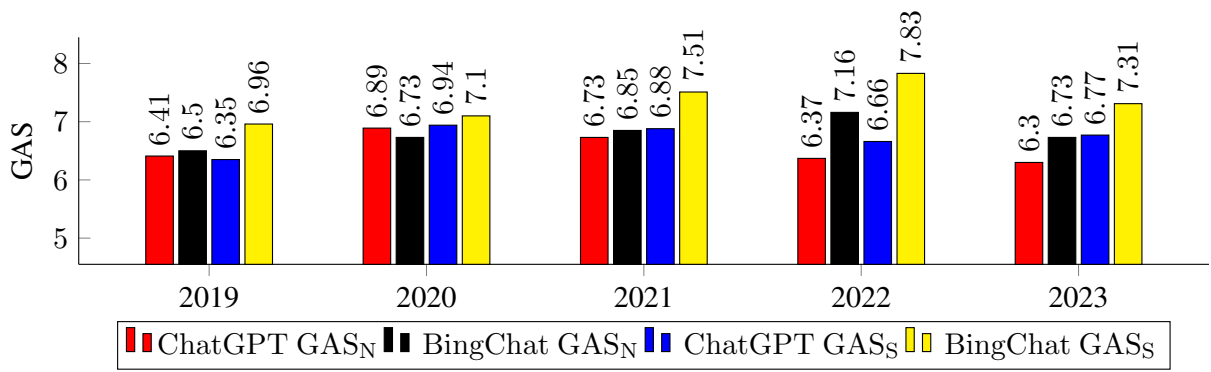
\begin{figure*}[h!]
		\begin{center}
			\resizebox{\textwidth}{!}{%
				\begin{tikzpicture}  
					\begin{axis}  
						[  
						ybar, 
						bar width=10pt,
						enlargelimits=0.15, 
						legend style={at={(0.5,-0.25)}, 	
							anchor=north,legend columns=-1},       
						ylabel={GAS}, 
						symbolic x coords={
							2019,
							2020,
							2021,
							2022,
							2023,
						},  
						xtick=data,  
						nodes near coords,  
						every node near coord/.append style={rotate=90, anchor=west},  
						width=\textwidth, 
						ymin=5,
						ymax=8,
						height=4.5cm, 
						width=16cm,
						axis x line*=bottom, axis y line*=left
						]  
						
						\addplot [fill=red] coordinates {
							(2019,6.41)
							(2020,6.89)
							(2021,6.73)
							(2022,6.37)
							(2023,6.30)						
						};
						\addplot [fill=black] coordinates { 
							(2019,6.50)
							(2020,6.73)
							(2021,6.85)
							(2022,7.16)
							(2023,6.73)						
						};
						\addplot [fill=blue] coordinates {
							(2019,6.35)
							(2020,6.94)
							(2021,6.88)
							(2022,6.66)
							(2023,6.77)
						};
						\addplot [fill=yellow] coordinates { 
							(2019,6.96)
							(2020,7.10)
							(2021,7.51)
							(2022,7.83)
							(2023,7.31)
						};
						\legend{$\mathrm{ChatGPT~GAS_N}$,~$\mathrm{BingChat~GAS_N}$,~$\mathrm{ChatGPT~GAS_S}$,~$\mathrm{BingChat ~GAS_S}$} 
					\end{axis}  
				\end{tikzpicture}
			}
		\end{center}
		\caption{Graph of combination scores for natural and social combinations from 2019 to 2023.}
		\label{fig:GAS}
	\end{figure*}
	
	The study conducted by~\cite{dao2023exa} showcased the ability of ChatGPT to attain a satisfactory performance level in the VNHSGE examination. This section focuses on the analysis of BingChat in the context of the VNHSGE examination. Table~\ref{tab:GAS} presents a comparative analysis of the performance of ChatGPT and BingChat in the domains of natural and social combination subjects over the period from 2019 to 2023. It is evident that ChatGPT consistently exhibits high performance in both natural and social contexts, without any notable disparity in scores. Nevertheless, the results of our research demonstrate that BingChat exhibits superior performance compared to ChatGPT in the domain of social combination. It is worth mentioning that BingChat exhibits comparable performance in natural combination and superior performance in social cobination when compared to ChatGPT, as depicted in Figure~\ref{fig:GAS}. As a result, successful completion of the VNHSGE examination guarantees the eligibility of BingChat to function as an educational virtual assistant tool. 
	
	\section{BingChat is better than ChatGPT for Vietnamese students}
	
	BingChat and ChatGPT are two language models (LLMs) that possess versatile applications across domains such as education, research, and entertainment. Nevertheless, BingChat possesses several notable advantages in comparison to ChatGPT, rendering it a more favorable option for Vietnamese students.
	
	\begin{itemize}
		
		\item \textbf{Availability in Vietnam}: BingChat is officially available in Vietnam, while ChatGPT is not. This implies that Vietnamese students have the ability to directly access BingChat without the need for a VPN or any other alternative method.
		
		\item \textbf{GPT-4 Technology}\footnote{\href{https://blogs.bing.com/search/march_2023/Confirmed-the-new-Bing-runs-on-OpenAI\%E2\%80\%99s-GPT-4}{https://blogs.bing.com/search/march\_2023/Confirmed-the-new-Bing-runs-on-OpenAI\%E2\%80\%99s-GPT-4}}: BingChat uses the more advanced GPT-4 technology, compared to ChatGPT, which is based on GPT-3.5. This means that BingChat is better able to understand and respond to complex queries, providing more accurate and informative answers.
		
		\item \textbf{Hyperlink Feature}: BingChat also offers a hyperlink feature that facilitates students in conveniently accessing supplementary resources and exploring related information. This particular attribute amplifies the educational encounter and furnishes students with a more extensive array of information that surpasses the responses generated by the model. (Refer to the Appendix, specifically Section~\ref{section:Hyperlink Feature}, for an illustrative example.)
		
		\item \textbf{Better Performance in Most Subjects}: BingChat demonstrates superior performance compared to ChatGPT across a range of academic disciplines, including mathematics, English, physics, chemistry, biology, history, geography, and civic education. The model's effectiveness in delivering precise responses and informative explanations in various subjects renders it a valuable resource for Vietnamese students seeking support in a diverse array of disciplines.
		
		\item \textbf{Access to Online Information}: BingChat is additionally linked to the Bing search engine, thereby granting it access to an extensive collection of publicly accessible online information. One notable benefit of BingChat is its ability to retrieve precise responses by leveraging online resources, thereby furnishing students with dependable and pertinent information. (Refer to the Appendix, specifically Section~\ref{section:Access to Online Information}, for an illustrative example.)
		
		\item \textbf{Real-time Data}: While Bing is primarily designed to offer up-to-date information from the internet, ChatGPT relies on the knowledge it has acquired through training, which has a specific cutoff date. This implies that ChatGPT's access to the most recent information or updates is limited to its training data, which encompasses information up until September 2021. (Refer to the Appendix, specifically Section~\ref{section:Hyperlink Feature}, for an illustrative example.)
		
	\end{itemize}
	
	In general, BingChat presents a range of benefits in comparison to ChatGPT. These advantages encompass its accessibility in Vietnam, implementation of GPT-4 technology, inclusion of hyperlinks functionality, notable proficiency across various subjects, provision of online information access, and ability to provide real-time data. The aforementioned benefits render BingChat a favored option among Vietnamese students in search of a proficient and dependable language model to fulfill their educational requirements.
	
	\section{Conclusion}
	
	This study presents an analysis indicating that Microsoft Bing Chat exhibits superior performance and capability compared to OpenAI ChatGPT specifically for Vietnamese students. BingChat, leveraging the advanced GPT-4 technology, demonstrates enhanced comprehension and agility in addressing intricate inquiries, resulting in the production of text that is both more precise and informative. While ChatGPT demonstrates exceptional performance in the field of literature, BingChat surpasses it in various other academic disciplines including mathematics, physics, chemistry, biology, history, geography, and civic education. The presence of this technology in Vietnam facilitates direct utilization of its functionalities, in contrast to the unavailability of ChatGPT within the country at present. Furthermore, the integration of hyperlinks within BingChat's responses enhances the educational process by offering supplementary materials and facilitating deeper investigation. In general, BingChat is identified as the superior option for Vietnamese students due to its advanced functionalities and improved accessibility.
	
	\bibliography{paper_7_9}
	\bibliographystyle{acl_natbib}
	
	\appendix
	
	\section{Hyperlink Feature}
	\label{section:Hyperlink Feature}
	
	\begin{mdframed}[linewidth=1pt,linecolor=red] 
		\textbf{Prompt}: Summarize the paper: Can chatgpt pass the vietnamese
		national high school graduation examination?
		arXiv preprint arXiv:2306.09170.
		
		\textbf{ChatGPT}: I'm sorry, but I couldn't find any information on the specific paper you mentioned. The paper you referenced, titled "Can ChatGPT Pass the Vietnamese National High School Graduation Examination?" and with the arXiv preprint number arXiv:2306.09170, does not exist in the current academic database I have access to. It's possible that the paper has not been widely published or is not available in the public domain at the moment. If you have any other questions or need assistance with a different topic, feel free to ask!
		
		\textbf{BingChat}: Sure! \href{https://arxiv.org/abs/2306.09170}{The paper titled “Can ChatGPT pass the Vietnamese National High School Graduation Examination?” by Xuan-Quy Dao, Ngoc-Bich Le, Xuan-Dung Phan, Bac-Bien Ngo highlights the potential of AI-powered chatbots in education and presents the results of using ChatGPT, a large language model, to complete the Vietnamese National High School Graduation Examination (VNHSGE)$^1$}.

		The study dataset included 30 essays in the literature test case and 1,700 multiple-choice questions designed for other subjects. \href{https://arxiv.org/abs/2306.09170}{The results showed that ChatGPT was able to pass the examination with an average score of 6-7, demonstrating the technology’s potential to revolutionize the educational landscape$^1$}.
		
		\href{https://arxiv.org/abs/2306.09170}{The analysis of ChatGPT performance revealed its proficiency in a range of subjects, including mathematics, English, physics, chemistry, biology, history, geography, civic education, and literature, which suggests its potential to provide effective support for learners$^1$}. \href{https://arxiv.org/abs/2306.09170}{However, further research is needed to assess ChatGPT performance on more complex exam questions and its potential to support learners in different contexts$^1$}. Is there anything else you would like to know about this paper? 
		
		Learn more:	1. \href{https://arxiv.org/abs/2306.09170}{arxiv.org}	2. \href{https://arxiv.org/abs/2306.09170}{arxiv.org}	3. \href{https://arxiv.org/abs/2306.09170}{arxiv.org}
		
		What is ChatGPT?	How does the Vietnamese National High School Graduation Examination work?
		
	\end{mdframed}

	\section{Access to Online Information}
	\label{section:Access to Online Information}
	
	\begin{mdframed}[linewidth=1pt,linecolor=red] 
		\textbf{Prompt}: Nơi nào sau đây ở nước ta có động đất biểu hiện rất yếu?
		A. Tây Bắc.
		B. Đông Bắc.
		C. Miền Trung.
		D. Nam Bộ.
		(Which of the following places in our country have very weak earthquakes?
		A. Northwest.
		B. Northeast.
		C. Central.
		D. Southern Vietnam.)
		
		\textbf{ChatGPT}: The answer to the question "Which of the following regions in our country has very weak seismic activity?" is option D. Nam Bộ.
		
		\textbf{BingChat}: Câu trả lời đúng cho câu hỏi này là D. Nam Bộ. Ở nước ta, Tây Bắc là khu vực có hoạt động động đất mạnh nhất, rồi đến khu vực Đông Bắc. \href{https://khoahoc.vietjack.com/question/89561/noi-nao-sau-day-o-nuoc-ta-co-dong-dat-bieu-hien-rat-yeu}{ Khu vực miền Trung ít động đất hơn, còn Nam Bộ biểu hiện động đất rất yếu$^1$}(The correct answer to this question is D. Nam Bo. In our country, the Northwest is the region with the strongest earthquake activity, followed by the Northeast. The Central region has fewer earthquakes, while the South shows very weak earthquakes).
		
		Learn more:	
		1. \href{https://khoahoc.vietjack.com/question/89561/noi-nao-sau-day-o-nuoc-ta-co-dong-dat-bieu-hien-rat-yeu}{khoahoc.vietjack.com}
		2. \href{https://tailieumoi.vn/cau-hoi/noi-nao-sau-day-o-nuoc-ta-co-dong-dat-bieu-hien-rat-yeu-parnk-580535.html}{tailieumoi.vn}	
		3. \href{https://tracnghiem.net/de-thi/cau-hoi-khu-vuc-co-hoat-dong-dong-dat-bieu-hien-rat-yeu-o-nuoc-ta-la-khu-vuc-nao-183514.html}{tracnghiem.net}
		4. \href{https://doctailieu.com/trac-nghiem/noi-nao-sau-day-o-nuoc-ta-co-dong-dat-bieu-hien-rat-yeu-a-tay-bac-trac-nghiem-90871}{doctailieu.com}
		5. \href{https://exam24h.com/question/noi-nao-sau-day-o-nuoc-ta-co-dong-dat-bieu}{exam24h.com}
		
		Câu 2 là gì (What is question 2)?  Tại sao Tây Bắc và Đông Bắc có hoạt động động đất mạnh hơn so với các khu vực khác? (Why do the Northwest and Northeast have stronger earthquake activity than other regions)?

	\end{mdframed}

\end{document}